\documentclass[11pt,a4paper]{article}
\usepackage[hyperref]{emnlp-ijcnlp-2019}
\usepackage{times}
\usepackage{latexsym}

\usepackage{url}
\usepackage{graphicx}
\usepackage{amsmath}
\usepackage{amsfonts}
\usepackage{multirow}
\usepackage{cleveref}
\usepackage{algorithm}
\usepackage{algorithmic}
\crefname{section}{§}{§§}

\aclfinalcopy % Uncomment this line for the final submission

\title{Core Semantic First: A Top-down Approach for AMR Parsing\thanks{~The work described in this paper is substantially supported by a grant from the Research Grant Council of the Hong Kong Special Administrative Region, China (Project Code: 14204418). The first author is grateful for the discussions with Zhisong Zhang and Zhijiang Guo.}}

\author{Deng Cai \\
	The Chinese University of Hong Kong\\
	{\tt thisisjcykcd@gmail.com} \\\And
	Wai Lam \\
	The Chinese University of Hong Kong\\
	{\tt wlam@se.cuhk.edu.hk} \\}

\date{}
\begin{document}
	\maketitle
	\begin{abstract}
		We introduce a novel scheme for parsing a piece of text into its Abstract Meaning Representation (AMR): Graph Spanning based Parsing (GSP). One novel characteristic of GSP is that it constructs a parse graph incrementally in a top-down fashion. Starting from the root, at each step, a new node and its connections to existing nodes will be jointly predicted. The output graph spans the nodes by the distance to the root, following the intuition of first grasping the main ideas then digging into more details. The \textit{core semantic first} principle emphasizes capturing the main ideas of a sentence, which is of great interest. We evaluate our model on the latest AMR sembank and achieve the state-of-the-art performance in the sense that no heuristic graph re-categorization is adopted. More importantly, the experiments show that our parser is especially good at obtaining the core semantics.
	\end{abstract}
	\section{Introduction}
	\label{intro}
\begin{figure}[t]
		\centering
		\includegraphics[scale=0.4]{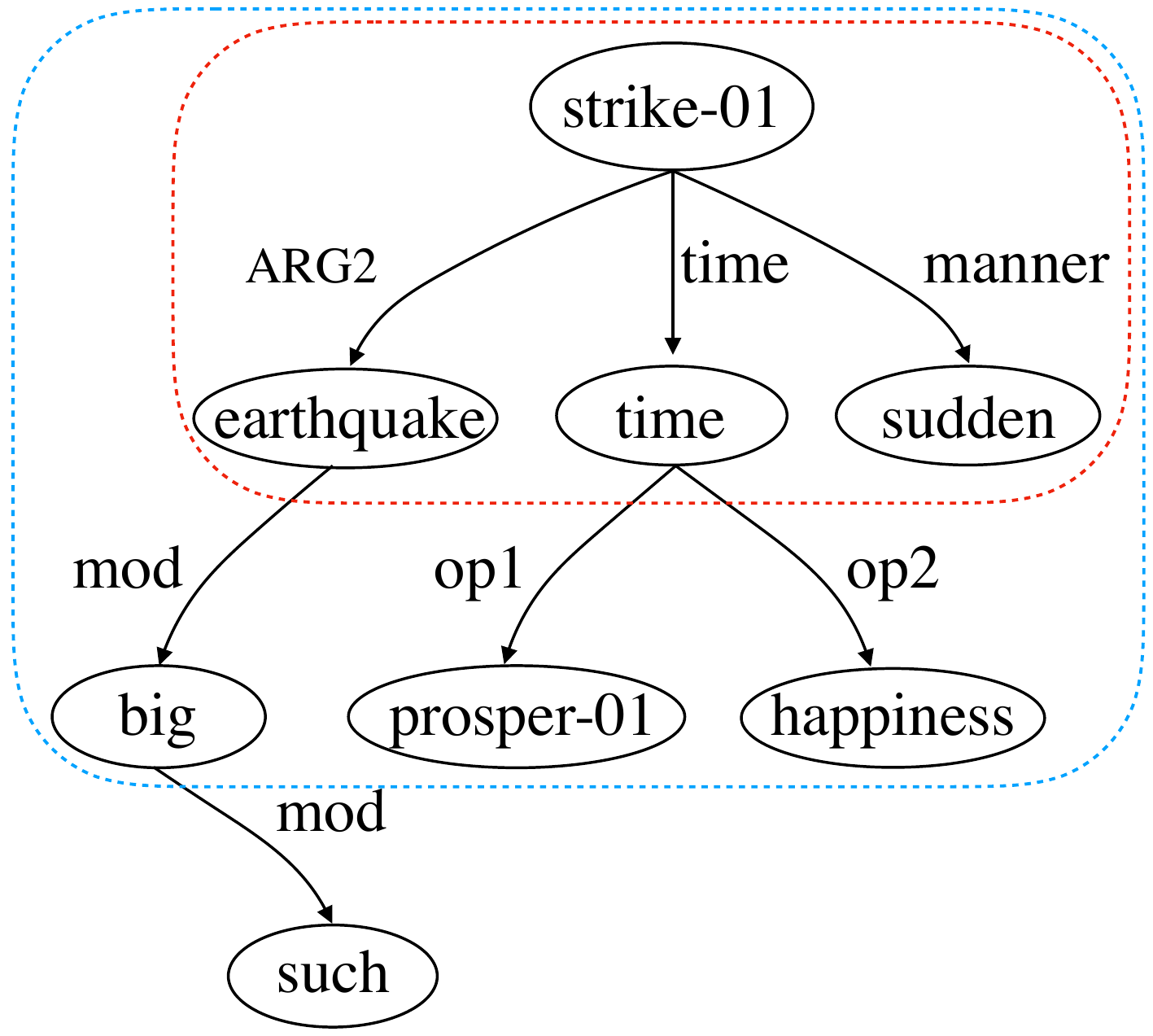}
		\caption{AMR for the sentence ``\textit{During a time of prosperity and happiness, such a big earthquake suddenly struck.}", where the subgraphs close to the root represent the core semantics.}
		\label{example}
	\end{figure}

	Abstract Meaning Representation (AMR) \cite{banarescu2013abstract} is a semantic formalism that encodes the meaning of a sentence as a rooted labeled directed graph. As illustrated by an example in Figure \ref{example},  AMR abstracts away from the surface forms in text, where the root serves as a rudimentary representation of the overall focus while the details are elaborated as the depth of the graph increases. AMR has been proved useful for many downstream NLP tasks, including text summarization \cite{liu-etal-2015-toward,hardy-vlachos-2018-guided} and question answering \cite{mitra2016addressing}.
	
	The task of AMR parsing is to map natural language strings to AMR semantic graphs automatically. Compared to constituent parsing \cite{zhang2009transition} and dependency parsing \cite{kubler2009dependency}, AMR parsing is considered more challenging due to the following characteristics: (1) The nodes in AMR have no explicit alignment to text tokens; (2) The graph structure is more complicated because of frequent reentrancies and non-projective arcs; (3) There is a large and sparse vocabulary of possible node types (concepts).
	
	Many methods for AMR parsing have been developed in the past years, which can be categorized into three main classes: Graph-based parsing \cite{flanigan2014discriminative,lyu2018amr} uses a pipeline design for concept identification and relation prediction. Transition-based parsing \cite{wang2016camr,damonte2016incremental,ballesteros2017amr,guo2018better,liu2018amr,wang2017getting} processes a sentence from left-to-right and constructs the graph incrementally. The third class is seq2seq-based parsing \cite{barzdins2016riga,konstas2017neural,van2017neural}, which views parsing as sequence-to-sequence transduction by a linearization (depth-first traversal) of the AMR graph. 
	
	While existing graph-based models cannot sufficiently model the interactions between individual decisions, the autoregressive nature of transition-based and seq2seq-based models makes them suffer from error propagation, where later decisions can easily go awry, especially given the complexity of AMR. Since capturing the core semantics of a sentence is arguably more important and useful in practice, it is desirable for a parser to have a global view and a priority for capturing the main ideas first. In fact, AMR graphs are organized in a hierarchy that the core semantics stay closely to the root, for which a top-down parsing scheme can fulfill the desiderata. For example, in Figure \ref{example}, the subgraph in the red box already conveys the core meaning ``\textit{an earthquake suddenly struck at a particular time}", and the subgraph in the blue box further informs that ``\textit{the earthquake was big}" and ``\textit{the time was of prosperity and happiness}".
	
	We propose a novel framework for AMR parsing known as Graph Spanning based Parsing (GSP). One novel characteristic of GSP is that, to our knowledge, it is the first top-down AMR parser.\footnote{Depth-first traversal in seq2seq models does not produce a strictly top-down order due to the reentrancies in AMR.} GSP performs parsing in an incremental, root-to-leaf fashion, but still maintains a global view of the sentence and the previously derived graph. At each step, it generates the connecting arcs between the existing nodes and the coming new node, upon which the type of the new node (concept) is jointly decided. The output graph spans the nodes by the distance to the root, following the intuition of first grasping the main ideas then digging into more details. Compared to previous graph-based methods, our model is capable of capturing more complicated intra-graph interactions, while reducing the number of parsing steps to be linear in the sentence length.\footnote{Since the size of AMR graph is approximately linear in the length of sentence.} Compared to transition-based methods, our model removes the left-to-right restriction and avoids sophisticated oracle design for handling the complexity of AMR graphs.
	
	Notably, most existing methods including the state-the-of-art parsers often rely on heavy graph re-categorization for reducing the complexity of the original AMR graphs. For graph re-categorization, specific subgraphs of AMR are grouped together and assigned to a single node with a new compound category \cite{werling2015robust,wang2017getting,foland2017abstract,lyu2018amr,groschwitz2018amr,guo2018better}. The hand-crafted rules for re-categorization are often non-trivial, requiring exhaustive screening and expert-level manual efforts. For instance, in the re-categorization system of \newcite{lyu2018amr}, the graph fragment ``$\texttt{temporal-quantity}\stackrel{:ARG3-of}{\longrightarrow}\texttt{rate-entity-91}\stackrel{:unit}{\longrightarrow}\texttt{year}\stackrel{:quant}{\longrightarrow}\texttt{1}$" will be replaced by one single nested node ``$\texttt{rate-entity-3(annual-01)}$". There are hundreds of such manual heuristic rules. This kind of re-categorization has been shown to have considerable effects on the performance \cite{wang2017getting,guo2018better}. However, one issue is that the precise set of re-categorization rules differs among different models, making it difficult to distinguish the performance improvement from model optimization or carefully designed rules. In fact, some work will become totally infeasible when removing this re-categorization step. For example, the parser of \newcite{lyu2018amr} requires tight integration with this step as it is built on the assumption that an injective alignment exists between sentence tokens and graph nodes.
	
	We evaluate our parser on the latest AMR sembank and achieve competitive results to the state-of-the-art models. The result is remarkable since our parser directly operates on the original AMR graphs and requires no manual efforts for graph re-categorization. The contributions of our work are summarized as follows:
	\begin{itemize}
		\item We propose a new method for learning AMR parsing that produces high-quality core semantics.
		\item Without the help of heuristic graph re-categorization which requires expensive expert-level manual efforts for designing re-categorization rules, our method achieves state-of-the-art performance.
	\end{itemize}
	\section{Related Work}
	\label{related}
	Currently, most AMR parsers can be categorized into three classes: (1) \textit{Graph-based} methods \cite{flanigan2014discriminative,flanigan2016cmu,werling2015robust,foland2017abstract,lyu2018amr,zhang-etal-2019-amr} adopt a pipeline approach for graph construction. It first maps continuous text spans into AMR concepts, then calculates the scores of possible edges and uses a maximum spanning connected subgraph algorithm to select the final graph. The major deficiency is that the concept identification and relation prediction are strictly performed in order, yet the interactions between them should benefit both sides \cite{zhou-etal-2016-amr}. In addition, for computational efficacy, usually only first-order information is considered for edge scoring.  (2) \textit{Transition-based} methods \cite{wang2016camr,damonte2016incremental,wang2017getting,ballesteros2017amr,liu2018amr,peng2018amr,guo2018better,naseem-etal-2019-rewarding} borrow techniques from shift-reduce dependency parsing. Yet the non-trivial nature of AMR graphs (e.g., reentrancies and non-projective arcs) makes the transition system even more complicated and difficult to train \cite{guo2018better}. (3) \textit{Seq2seq-based} methods \cite{barzdins2016riga,peng-etal-2017-addressing,konstas2017neural,van2017neural} treat AMR parsing as sequence-to-sequence problem by linearizing AMR graphs, thus existing seq2seq models \cite{bahdanau2014neural,luong-pham-manning:2015:EMNLP} can be readily utilized. Despite its simplicity, the performance of the current seq2seq models lag behind when the training data is limited. The first reason is that seq2seq models are often not as effective on smaller datasets. The second reason is that the linearized AMRs add the challenges of making use of the graph structure information.
	
	There are also some notable exceptions. \newcite{peng2015synchronous} introduce a synchronous hyperedge replacement grammar solution. \newcite{pust2015parsing} regard the task as a machine translation problem, while \newcite{artzi2015broad} adapt combinatory categorical grammar. \newcite{groschwitz2018amr,lindemann-etal-2019-compositional} view AMR graphs as the structure AM algebra.
	
	Most AMR parsers require an explicit alignment between tokens in the sentences and nodes in the AMR graph during training. Since such information is not annotated, a pre-trained aligner \cite{flanigan2014discriminative,pourdamghani2014aligning,liu2018amr}  is often required. More recently, \newcite{lyu2018amr} demonstrate that the alignments can be treated as latent variables in a joint probabilistic model.
	\section{Background and Overview}
	\label{overview}
	\subsection{Background of Multi-head Attention}
	The multi-head attention mechanism introduced by \newcite{vaswani2017attention} is used as a basic building block in our framework. The multi-head attention consists of $H$ attention heads, and each of which learns a distinct attention function. Given a query vector $x$ and a set of vectors $\{y_1, y_2, \ldots, y_m\}$ or in short $y_{1:m}$, for each attention head, we project $x$ and $y_{1:m}$ into distinct query, key, and value representations $q \in \mathbb{R}^d$, $K \in \mathbb{R}^{m\times d}$ and $V \in \mathbb{R}^{m\times d}$ respectively, where $d$ is the dimension of the vector space. Then we perform scaled dot-product attention \cite{vaswani2017attention}:
	\begin{align*}
	&a = \text{softmax}\frac{(Kq)}{\sqrt{d}} \\
	&attn = a V
	\end{align*}
	where $a \in \mathbb{R}^m$ is the attention vector (a distribution over all input $y_{1:m}$) and $attn$ is the weighted sum of the value vectors. Finally, the outputs of all attention heads are concatenated and projected to the original dimension of $x$. For brevity, we will denote the whole attention procedure described above as a function $T(x,y_{1:m})$.
	
	Based on the multi-head attention, the Transformer encoder \cite{vaswani2017attention} uses self-attention for context information aggregation when given a set of vectors (e.g., word embeddings in a sentence or node embeddings in a graph).
	\subsection{Overview}
	\begin{figure*}[t]
		\centering
		\includegraphics[scale=0.45]{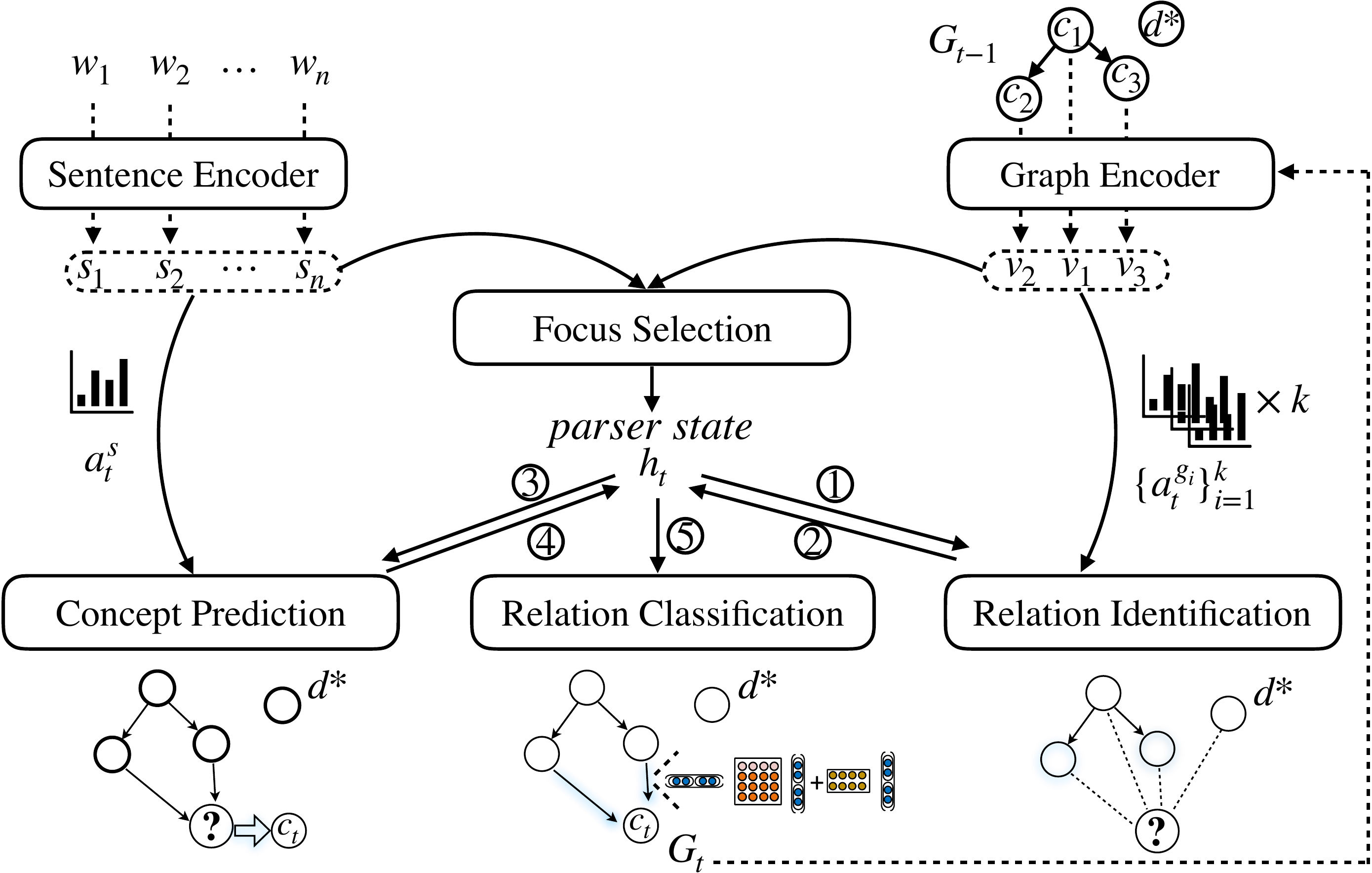}
		\caption{Model architecture of GSP, together with the decoding procedure at the time step $t$, where the read and write operations around the parser state $h_t$ follow the order $\textcircled{1}\rightarrow\textcircled{2}\rightarrow\textcircled{3}\rightarrow\textcircled{4}\rightarrow\textcircled{5}$.}
		\label{arch}
	\end{figure*}
	Figure \ref{arch} depicts the major neural components in our proposed framework: The Sentence Encoder component and the Graph Encoder component are designed for token-level sentence representation and node-level graph representation respectively. Given an input sentence $\mathbf{w} = (w_1, w_2, \ldots, w_n)$, where $n$ is the sentence length, the Sentence Encoder component will first read the whole sentence and encode each word $w_i$ into the hidden state $s_i$. The initial graph $G_0$ is always initialized with one dummy node $d^*$ and a previously generated concept $c_j$ is encoded into the hidden state $v_j$ by the Graph Encoder component.
	
	At each time step $t$, the Focus Selection component reads both the sentence representation $s_{1:n}$ and the graph representation $v_{0:t-1}$ of $G_{t-1}$ repeatedly, generates the initial parser state $h_t$. The parser state carries the most useful information and serves as a writable memory during the expansion step. Next, the Relation Identification component decides which specific head nodes to expand by computing the multiple attention scores $\{a^{g_i}_t\}_{i=1}^k$ over the existing nodes. New arcs are generated according to the attention scores. Then the Concept Prediction component updates the parser state $h_t$ with arc information, computes the attention vector $a^s_t$ over the sentence and accordingly chooses a specific part to generate the new concept $c_t$. Finally, the Relation Classification component is used to predict the relation labels between the newly generated concept and its predecessors. Consequently an updated graph $G_t$ is produced and $G_t$ will be processed for the next time step. The whole decoding procedure is terminated if the newly generated concept is the special stop concept $\oslash$.
	
	Our method expands the graph in a root-to-leaf fashion, nodes with shorter distances to the root will be introduced first. It follows a similar way that humans grasp the meaning: first seeking the main concepts then proceeding to the sub-structures governed by certain head concepts \cite{banarescu2013abstract}.
	
	During training, we use breadth-first search to decide the order of nodes. However, for nodes with multiple children, there still exist multiple valid selections. In order to define a deterministic decoding process, we sort sibling nodes by their relations to the head node. We will present more discussions on the choice of sibling order in \cref{sibling}.
	\section{Framework Description}
	\subsection{Sentence \& Graph Representation}
	Transformer encoder architecture is employed for both the Sentence Encoder and the Graph encoder components. For sentence encoding, a special token ($\rhd$) is prepended to the input word sequence, whose final hidden state $s_0$ is regarded as an aggregated summary of the whole sentence and used as the initial state in parsing steps.
	
	The Graph Encoder component takes previously generated concept sequence $(c_0, c_1, \ldots, c_{t-1})$ ($c_0$ is the dummy node $d^*$) as input. For computation efficiency and reducing error propagation, instead of encoding the edge information explicitly, we use the Transformer encoder to capture the interactions between nodes. Finally, the encoder outputs a sequence of node representations $(v_0, v_1, \dots, v_{t-1})$.
	\subsection{Focus Selection}
	\label{rr}
	At each time step $t$, the Focus Selection component will read the sentence and the partially constructed graph repeatedly for gradually locating and collecting the most relevant information for the next expansion. We simulate the repeated reading by multiple levels of attention. Formally, the following recurrence is applied by $L$ times:
	\begin{align*}
	&x^{(l+1),1}_t =\text{LN}( h^{(l)}_t+ T^{(l+1),1}(h^{(l)}_t, s_{1:n})) \\
	&x^{(l+1),2}_t = \text{LN}(x^{(l+1),1}_t+T^{(l+1), 2}(x^{(l+1),1}_t, v_{0:t-1}) )\\
	&h^{(l+1)}_t = \max(x^{(l+1),2}_t W^{l+1}_1 +b^{l+1}_1) W^{l+1}_2+b^{l+1}_2
	\end{align*}
	where $T(\cdot,\cdot)$ is the multi-head attention function. $\text{LN}$ is the layer normalization \cite{lei2016layer} and $h^{(0)}_t$ is always initialized with $s_0$. For clarity, we denote the last hidden state $h^{(L)}_t$ as $h_t$, as the parser state at the time step $t$. We now proceed to present the details of each decision stage of one parsing step, which is also illustrated in Figure \ref{arch}.
	\subsection{Relation Identification}
	\label{ri}
	Our Relation Identification component is inspired by a recent attempt of exposing auxiliary supervision on attention mechanism \cite{strubell-etal-2018-linguistically}. It can be considered as another attention layer over the existing graph, yet the attention weights explicitly indicate the likelihood of the new node being attached to a specific node. In other words, its aim is to answer the question of where to expand. Since a node can be attached to multiple nodes by playing different semantic roles, we utilize multi-head attention and take the maximum over different heads as the final arc probabilities.
	
	Formally, through a multi-head attention mechanism taking $h_t$ and $v_{0:t-1}$ as input, we obtain a set of attention weights $\{a^{g_i}_t\}_{i=1}^k$, where $k$ is the number of attention heads and $a^{g_i}_t$ is the $i$-th probability vector. The probability of the arc between the new node and the node $v_j$ is then computed by  $a^{g}_{t,j} = max_i(a^{g_i}_{t,j})$. Intuitively, each head is in charge of a set of possible relations (though not explicitly specified). If certain relations do not exist between the new node and any existing node, the probability mass will be assigned to the dummy node $d^*$. The maximum pooling reflects that the arc should be built once one relation is activated.\footnote{We also found that there may exist more than one relation between two distinct nodes, however, it rarely happens.}
	
	The attention mechanism passes the arc decisions to later layers by the update of the parser state as follows:
	\begin{equation}
	h_t = LN(h_t + W^{\text{arc}}\sum_{j=0}^{t-1} a^{g}_{t,j} v_j )
	\nonumber
	\end{equation}
	\subsection{Concept Prediction}
	\label{cp}
	Our Concept Prediction component uses a soft alignment between words and the new concept. Concretely, a single-head attention $a^s_t$ is computed based on the parser state $h_t$  and the sentence representation $s_{1:n}$, where $a^s_{t,i}$ denotes the attention weight of the word $w_i$ in the current time step. This component then updates the parser state with the alignment information via the following equation:
	\begin{equation}
	h_t = LN(h_t + W^{\text{conc}}\sum_{i=1}^{n} a^s_{t,i} s_i )
	\nonumber
	\end{equation}
	
	The probability of generating a specific concept $c$ from the concept vocabulary $\mathcal{V}$ is calculated as $gen(c|h_t) = {\exp({x_c}^Th_t)} / { \sum_{c'\in \mathcal{V}} \exp({x_{c'}}^Th_t)}$, where $x_c$ (for $c\in\mathcal{V}$) denotes the model parameters. To address the data sparsity issue in concept prediction, we introduce a copy mechanism in similar spirit to \newcite{gu2016incorporating}. Besides generation, our model can either directly copy an input token $w_i$ (e.g, for entity names) or map $w_i$ to one concept $m(w_i)$ according to the alignment statistics\footnote{Based on the alignments provided by \newcite{liu2018amr}, for each word, the most frequently aligned concept (or its lemma if it has empty alignment) is used for direct mapping.} in the training data (e.g., for ``went", it would propose \textit{go}). Formally, the prediction probability of a concept $c$ is given by:
	\begin{align*}
	P(c|h_t) = &P(copy|h_t) \sum_{i=1}^na^s_{t,i}[[w_i=c]] \\
	+& P(map|h_t)\sum_{i=1}^na^s_{t,i}[[m(w_i)=c]] \\ 
	+& P(gen|h_t) gen(c|h_t)
	\end{align*} 
	where $[[\ldots]]$ is the indicator function. $P(copy|h_t)$, $P(map|h_t)$ and $P(gen|h_t)$ are the probabilities of three prediction modes respectively, computed by a single layer neural network with softmax activation.
	\subsection{Relation Classification}
	\label{rc}
	Lastly, the Relation Classification component employs a multi-class classifier for labeling the arcs detected in the Relation Identification component. The classifier uses a biaffine function to score each label, given the head concept representation $v_i$ and the child vector $h_t$ as input:
	\begin{equation}
	e_t^i = h_t^TWv_i + U^Th_t + V^Tv_i + b
	\nonumber
	\end{equation}
	where $W, U, V, b$ are model parameters. As suggested by \newcite{dozat2016deep}, we project $v_i$ and $h_t$ to a lower dimension for reducing the computation cost and avoiding the overfitting of the model. The label probabilities are computed by a softmax function over all label scores.
	\subsection{Reentrancies}
    \noindent AMR reentrancy is employed when a node participates in multiple semantic relations (with multiple parent nodes), and that is why AMRs are graphs, rather than trees. The reentrancies are often hard to treat. While previous work often either remove them \cite{guo2018better} or relies on rule-based restoration in the postprocessing stage \cite{lyu2018amr,van2017neural}, our model provides a new and principled way to deal with reentrancies. In our approach, when a new node is generated, all its connections to already existing nodes are determined by the multi-head attention. For example, for a node with $k$ parent nodes, $k$ different heads will point to the those parent nodes respectively. For a better understanding of our model, a pseudocode is presented in Algorithm \ref{algo}.
    	 \begin{algorithm}[t]
    	\caption{Graph Spanning based Parsing}
    	\begin{algorithmic}[1]
    		\REQUIRE the input sentence $\mathbf{w}= (w_1, w_2, \ldots, w_n)$
    		\ENSURE the AMR graph $G$ corresponds to $\mathbf{w}$.\\
    		$\Diamond$ Learning Sentence Representation
    		\STATE \textbf{w} = $(w_0 = \rhd) + (w_1, w_2, \ldots, w_n)$
    		\STATE $s_0, s_1, s_2, \ldots, s_n = $ Transformer(\textbf{w})\\
    		$\Diamond$ Initialization
    		\STATE initialize the graph $G_0$ ($c_0=d^*$)
    		\STATE initialize time step $t =1$\\
    		$\Diamond$ Entering Main Spanning Loop
    		\WHILE {True}
    		\STATE $h_0, \ldots, v_{t-1}$ = Transformer($c_{0:t-1}$)
    		\STATE $h_t$ = Focus$\_$Selection $(s_0, v_{0:{t-1}}, s_{1:n})$
    		\STATE $h_t$ = Relation$\_$Identification $(h_t, v_{0:{t-1}})$
    		\item \textit{decide the parents nodes} $pred(t)$ \textit{of} $c_t$
    		\STATE $h_t$ = Concept$\_$Prediction $(h_t, s_{1:n})$
    		\item \textit{decide the node type of }$c_t$
    		\IF {$c_t == \oslash$ }
    		\STATE \textbf{break}
    		\ENDIF
    		\FOR{ $i\in pred(t)$}
    		\STATE Relation$\_$Classification $(h_t, v_i)$
    		\item \textit{decide the edge type between} $c_t$ \textit{and} $c_i$
    		\ENDFOR
    		\STATE update $G_{t-1}$ to $G_t$
    		\ENDWHILE
    		\RETURN $G_{t-1}$
    	\end{algorithmic}
    	\label{algo}
    \end{algorithm}
	\subsection{Training and Inference}
	Our model is trained to maximize the log likelihood of the gold AMR graphs given sentences, i.e. $\log P(G|\mathbf{w})$, which can be factorized as:
	\begin{align*}
	\log P(G|\mathbf{w}) &= \sum_{t=1}^m \bigg( \log P(c_t|G_{t-1}, \mathbf{w})  \\
	&+ \sum_{i\in pred(t)}\log P(arc_{it}|G_{t-1}, \mathbf{w}) \\
	& + \sum_{i\in pred(t)}\log P(rel_{arc_{it}}|G_{t-1}, \mathbf{w})\bigg)
	\end{align*}
	where $m$ is the total number of vertices. The set of predecessor nodes of $c_t$ is denoted as $pred(t)$. $arc_{it}$ denotes the arc between $c_i$ and $c_t$, and $rel_{arc_{it}}$ indicates the arc label (relation type).
	
	As mentioned, GSP is an autoregressive model, such as seq2seq models and transition models, but it factors the distribution according to a top-down graph structure rather than a depth-first traversal or a left-to-right chain. Meanwhile, GSP has a clear separation of node, arc and relation label probabilities, interacting in a more interpretable and tighten manner.
	
	At the operational or testing time, the prediction for the input $\mathbf{w}$ is obtained via $\hat{G} = \arg\max_{G'} P(G'|\mathbf{w})$. Rather than iterating over all possible graphs, we adopt a beam search to approximate the best graph. Specifically, for each partially constructed graph, we only consider the top-$K$ concepts obtaining the best single-step probability (a product of the corresponding concept, arc, and relation label probability), where $K$ is the beam size. Only the best $K$ graphs at each time step are kept for the next expansion.
	\section{Experiments}
	\subsection{Setup}
	We focus on the most recent LDC2017T10 dataset, as it is the largest AMR corpus. It consists of 36521, 1368, and 1371 sentences in the training, development, and testing sets respectively.
	
	We use Stanford CoreNLP \cite{manning2014stanford} for text preprocessing, including tokenization, lemmatization, part-of-speech, and named-entity tagging. The input for sentence encoder consists of the randomly initialized lemma, part-of-speech tag, and named-entity tag embeddings, as well as the output from a learnable CNN with character embeddings as inputs. The graph encoder uses randomly initialized concept embeddings and another char-level CNN. Model hyper-parameters are chosen by experiments on the development set. The details of the hyper-parameter settings are provided in the Appendix. During testing, we use a beam size of $8$ for generating graphs.\footnote{Our code can be found at \url{https://github.com/jcyk/AMR-parser}.}

\begin{figure}[t]
	\centering
	\includegraphics[scale=0.20]{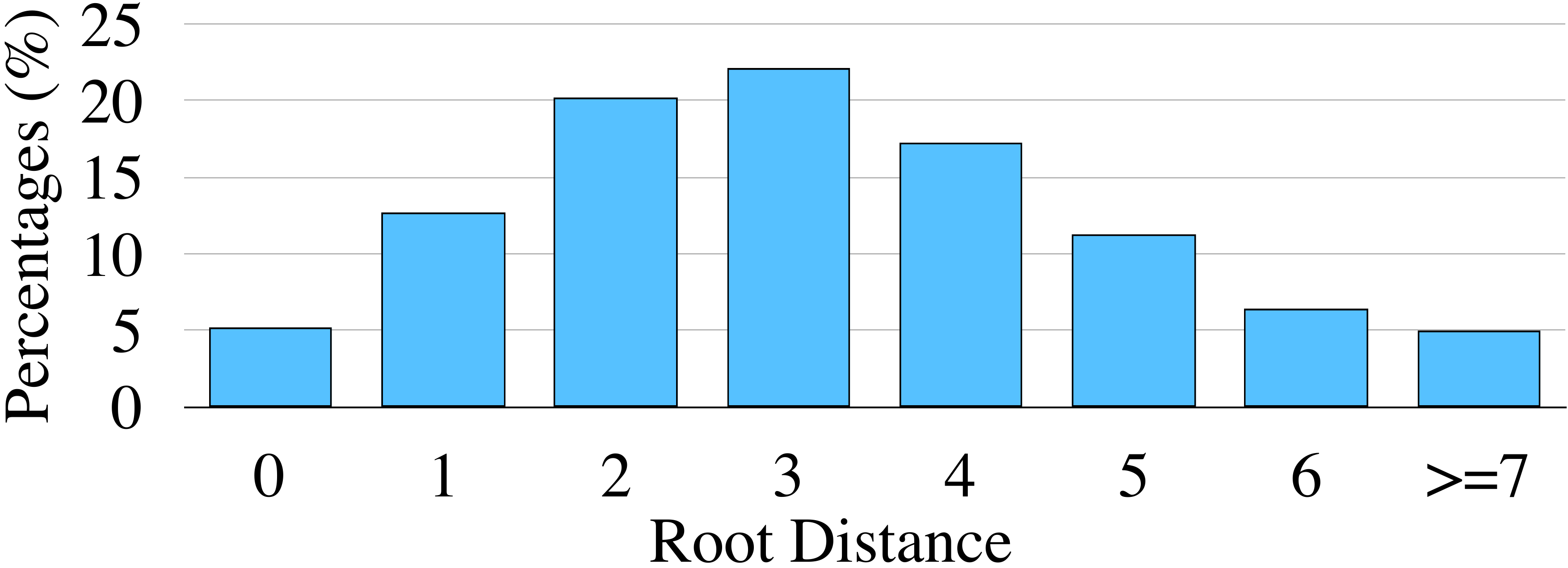}
	\caption{The distribution of root distance of concepts in the test set.}
	\label{dist}
\end{figure}

	Conventionally, the quality of AMR parsing results is evaluated using the Smatch tool \cite{cai2013smatch}, which seeks for the maximum number of overlaps between two AMR annotations after decomposing AMR graphs into triples. However, the ordinary Smatch metric treats all triples equally regardless of their roles in the composition of the whole sentence meaning. We refine the ordinary Smatch metric to take into consideration the notion of core semantics. Specifically, we compute:
	\begin{itemize}
		\item Smatch-weighted: This metric weights different triples by their importance of composing the core ideas. The root distance $d$ of a triple is defined as the minimum root distance of its involving nodes, the weight of the triple is then computed as:
		\begin{equation}
		w = \min(-d + d_{thr}, 1)
		\nonumber
		\end{equation}
		In other words, the weight has a linear decay in root distance until $d_{thr}$. If two triples are matched, the minimum importance score of them is obtained. In our experiments, $d_{thr}$ is set to 5. 
		\item Smatch-core: This metric only compares the subgraphs representing the main meaning. Precisely, we cut down AMR graphs by setting a maximum root distance $d_{max}$ and only keep the nodes and edges within the threshold. $d_{max}$ is set to $4$ in our experiments, of which the remaining subgraphs still have a broad coverage of the original meaning, as illustrated by the distribution of root distance in Figure \ref{dist}.
	\end{itemize}
	Besides, we also evaluate the quality by computing the following metrics.
	\begin{itemize}
		\item complete-match (CM): This metric counts the number of parsing results that are completely correct.
		\item root-accuracy (RA): This metric measures the accuracy of the root concept identification.
	\end{itemize}
\begin{table*}[t]
	\centering
	\begin{tabular}{c|c|c|c|c|c|c}
		\hline
		\multirow{2}{*}{Model} & Graph &\multicolumn{3}{|c|}{Smatch(\%)} & \multirow{2}{*}{RA(\%)} & \multirow{2}{*}{CM(\%)} \\
		\cline{3-5}
		&Re-ca.&weighted&core&ordinary&&\\
		\hline
		\newcite{buys2017oxford} &No &-&-&61.9&-&-\\
		\newcite{van2017neural} + 100K & No&68.8&67.6& 71.0&75.8&10.2\\
		\newcite{guo2018better} & Yes&63.5&62.3&69.8&63.6& 9.4\\
		\newcite{lyu2018amr} & Yes&66.6&67.1 & \textbf{74.4}&59.1&10.2\\
		\newcite{groschwitz2018amr} & Yes&-&-&71.0&-&-\\
		\hline
		Ours & No &\textbf{71.3}&\textbf{70.2} &73.2&\textbf{76.9}&\textbf{11.6}\\
		\hline
	\end{tabular}
	\caption{Comparison with state-of-the-art methods (results on the test set). Results relying on heuristic rules for graph re-categorization are marked ``Yes" in the Graph Re-ca. column.}
	\label{main_res}
\end{table*}
\begin{figure*}[t]
	\centering
	\includegraphics[scale=0.58]{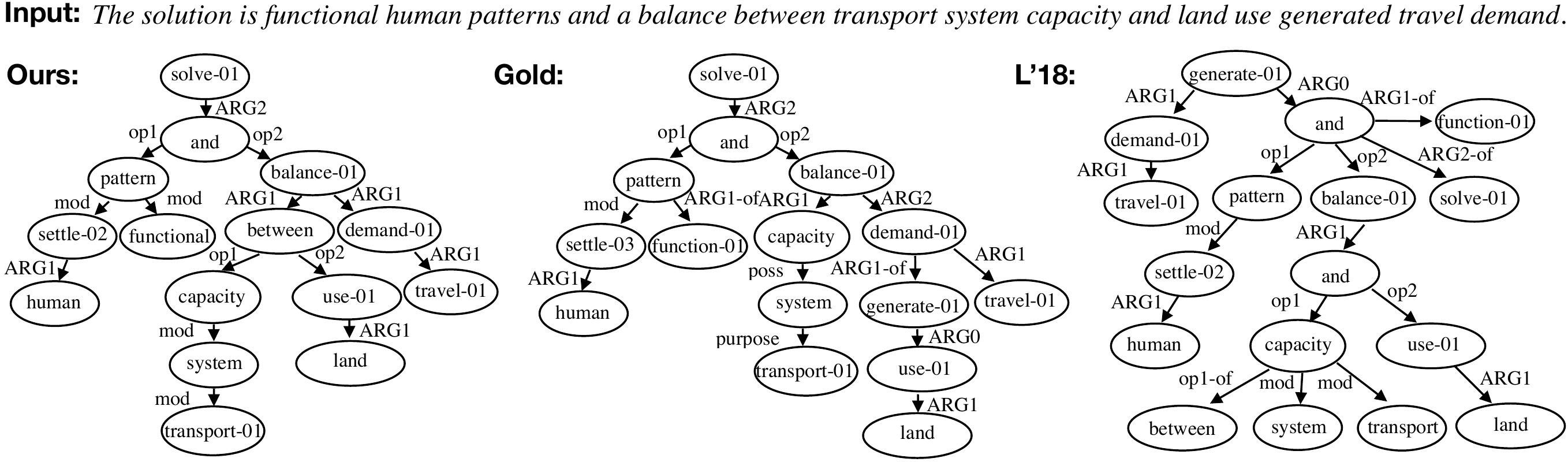}
	\caption{Case study.}
	\label{case}
\end{figure*}
	\subsection{Main Results and Case Study}
	The main result is presented in Table \ref{main_res}. We compare our method with the best-performing models in each category as discussed in \cref{related}.
	
	Concretely, \newcite{van2017neural} is a character-level seq2seq model that achieves very competitive result. However, their model is very data demanding as it requires to train on additional 100K sentence-AMR pairs generated by other parsers. \newcite{guo2018better} is a transition-based parser with refined search space for AMR. Certain concepts and relations (e.g., reentrancies) are removed to reduce the burdens during training. \newcite{lyu2018amr} is a graph-based method that achieves the best-reported result evaluated by the ordinary Smatch metric. Their parser uses different LSTMs for concept prediction, relation identification, and root identification sequentially. Also, the relation identification stage has the time complexity of $O(m^2\log m)$ where $m$ is the number of concepts. \newcite{groschwitz2018amr} views AMR as terms of the AM algebra \cite{groschwitz-etal-2017-constrained}, which allows standard tree-based parsing techniques to be applicable. The complexity of their projective decoder is $O(m^5)$. Last but not least, all these models except for that of \newcite{van2017neural} require hand-crafted heuristics for graph re-categorization.
	
	We consider the Smatch-weighted metric as the most suitable metric for measuring the parser's quality on capturing core semantics. The comparison shows that our method significantly outperforms all other methods. The Smatch-core metric also demonstrates the advantage of our method in capturing the core ideas. Besides, our model achieves the highest root-accuracy (RA) and complete-match (CM), which further confirms the usefulness of a global view and the core semantic first principle.
	
	Even evaluated by the ordinary Smatch metric, our model yields better results than all previously reported models with the exception of \newcite{lyu2018amr}, which relies on a tremendous amount of manual heuristics for designing rules for graph re-categorization and adopts a pipeline approach. Note that our parser constructs the AMR graph in an end-to-end fashion with a better (quadratic) time complexity.
	
	We present a case study in Figure \ref{case} with comparison to the output of \newcite{lyu2018amr}'s parser. As seen, both parsers make some mistakes. Specifically, our method fails to identify the concept \texttt{generated-01}. While \newcite{lyu2018amr}'s parser successfully identifies it, their parser mistakenly treats it as the root of the whole AMR. It leads to a serious drawback of making the sentence meaning be interpreted in a wrong way. In contrast, our method shows a strong capacity in capturing the main idea ``\textit{the solution is about some patterns and a balance}". However, on the ordinary Smatch metric, their graph obtains a higher score (68\% vs. 66\%), which indicates that the ordinary Smatch is not a proper metric for evaluating the quality of capturing core semantics. If we adopt the Smatch-weighted metric, our method achieves a better score i.e. 74\% vs. 61\%.
	\begin{figure}[t]
		\centering
		\includegraphics[scale=0.22]{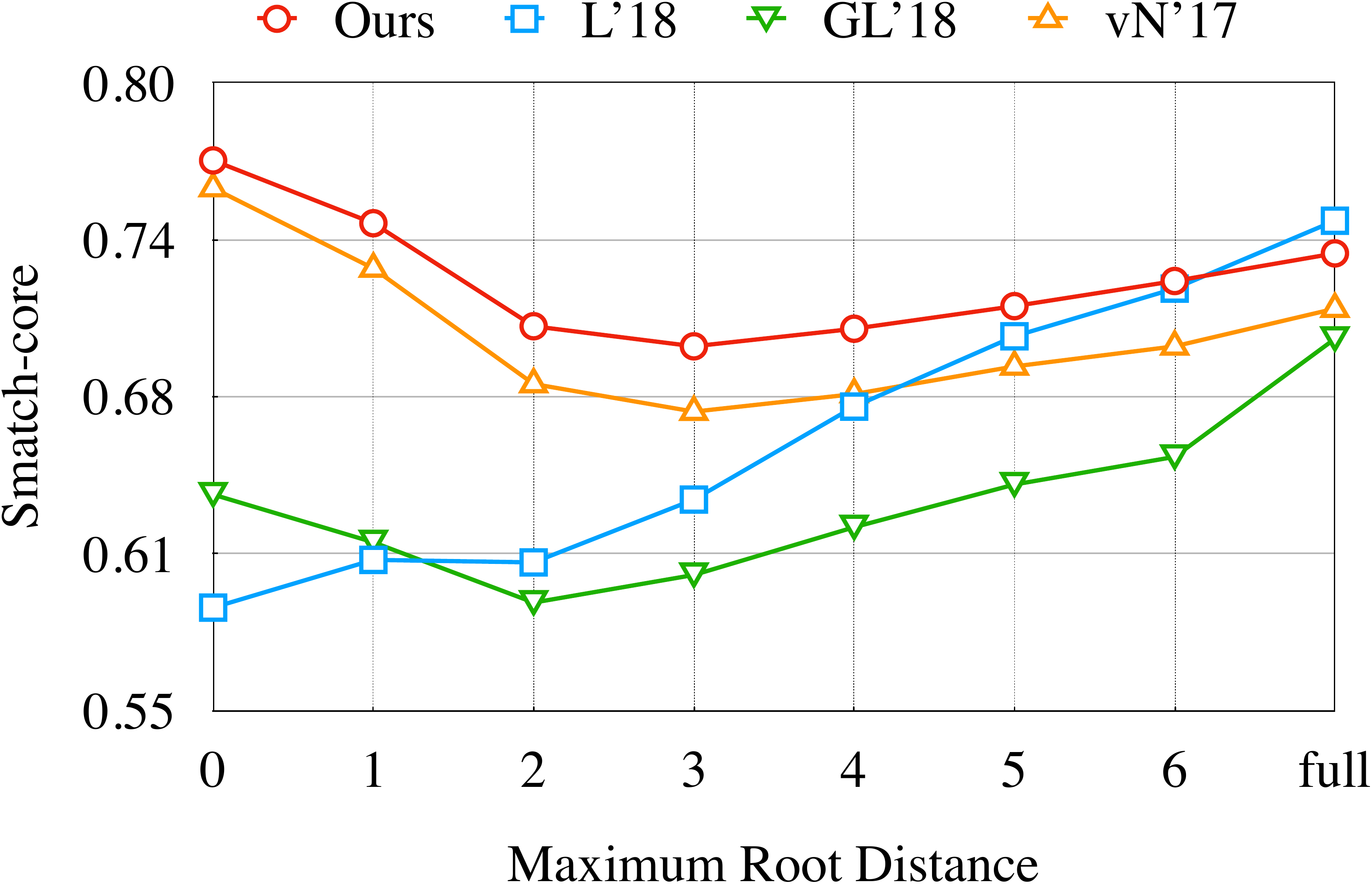}
		\caption{Smatch scores with different root distances. vN’17 is \newcite{van2017neural}'s parser with 100K additional training pairs. GL'18 is \newcite{guo2018better}'s parser. L'18 is \newcite{lyu2018amr}'s parser.}
		\label{root}
	\end{figure}
	\subsection{More Results}
	\label{sibling}
	To reveal our parser's ability for grasping meanings at different levels of granularity, we plot the Smatch-core scores in Figure \ref{root}  by varying the maximum root distance $d_{max}$, compared with several strong baselines and the state-of-the-art model. It demonstrates that our method is better at abstracting the core ideas of a sentence.
%	
%	In order to investigate how our parser performs on some additional aspects, we also use the fine-grained evaluation tool \cite{damonte2016incremental} and compare to systems which reported these scores. The results are shown in Table \ref{main_breakdown}. We obtain competitive results for all categories. Especially, we achieve the best results on reentrancies and negations, which are the most challenging sub-tasks that reflect the complexity of AMR graph. The comparison with \newcite{lyu2018amr} also provides some insights. Their method gives significantly better results on named entity recognition and the related wikification. The difference may be owing to their graph re-categorization that spends a lot of manual efforts on simplifying the entity graph structure.
	
	As discussed in \cref{overview}, there could be multiple valid generation orders for sibling nodes in an AMR graph. We experiment with the following traversal variants: (1) \textit{random}, which sorts the sibling nodes in completely random order. (2) \textit{relation freq.}, which sorts the sibling nodes according to their relations to the head node. We assign higher priorities to relations that occur more frequently, which drives our parser always to seek for the most common relation first. (3) \textit{combined}, which combines the above two strategies by using \textit{random} and \textit{relation freq.} with equal chance. As seen in Table \ref{variants}, the deterministic order strategy for training (\textit{relation freq.}) achieves better performance than random order. Interestingly,  the combined strategy significantly boosts the performance.\footnote{We note there are many other ways to generate a deterministic order. For example, \newcite{van2017neural} uses the order of aligned words in the sentence. However, we use the relation frequency method for its simplicity and not relying on external resources (e.g, an aligner). } The reason is that the random order potentially produces a larger set of training pairs since each random order strategy can be considered as a different training pair. On the other hand, the deterministic order stabilizes the maximum likelihood estimate training. Therefore, the combined strategy benefits from both worlds.
%	\begin{table}[t]
%		\centering
%		\begin{tabular}{c|c|c|c|c}
%			\hline
%			Metric &vN'17& GL'18 &L'18  & Ours  \\
%			\hline
%			Reentrancies &52&49&52&\textbf{55}\\
%			Concepts      &82&84&\textbf{86}& 84\\
%			Named Ent.   &79&80&\textbf{86}& 82\\
%			Wikification   &65&70&\textbf{76}& 73\\
%			Negations     &62&48&58& \textbf{63}\\
%			SRL               &66&63&\textbf{70}&67\\
%			\hline
%		\end{tabular}
%		\caption{Detailed results on various aspects.}
%		\label{main_breakdown}
%	\end{table}
	\section{Conclusion and Future Work}
	We presented the first top-down AMR parser. Our proposed parser builds a AMR graph incrementally in a root-to-leaf manner. Experiments show that our method has a better capability of capturing the core semantics in a sentence compared with previous state-of-the-art methods. In addition, we overcome the need of heuristics for graph re-categorization employed in most previous work, which makes our method much more transferable to other semantic representations or languages.
	
	Our methods follows the intuition that humans tend to grasp the core meaning of a sentence first. However, some cognitive theories \cite{langacker2008cognitive} also suggest that human language understanding is often presented as a circular, abductive process (hermeneutic circle). It is interesting to explore the use of some revision mechanisms when the initial steps go wrong.
	\begin{table}[t]
		\centering
		\begin{tabular}{c|c|c|c}
			\hline
			\multirow{2}{*}{Order} & \multicolumn{3}{c}{Smatch}\\
			\cline{2-4}
			& weighted & core & ordinary \\
			\hline
			random & 68.2 & 67.4& 70.4\\
			relation freq.& 69.9& 68.3&70.9 \\
			combined &71.3&70.2&73.2 \\
			\hline
		\end{tabular}
		\caption{The effect of different sibling orders. }
		\label{variants}
	\end{table}
	\bibliography{emnlp-ijcnlp-2019}
	\bibliographystyle{acl_natbib}
\appendix
\newpage
\section{Implementation Details}
In all experiments, we use the same char-level CNN settings in the sentence encoder and the graph encoder. In addition, all Transformer \cite{vaswani2017attention} layers in our model share the same hyper-parameter settings. For computation efficiency, we only allow each concept to attend to its previously generated concepts in the graph encoder.\footnote{Otherwise, we will need to re-compute the hidden states for all existing nodes at each parsing step.} Table \ref{setting} summarizes the chosen hyper-parameters after we tuned on the development set. To mitigate overfitting, we also apply dropout \cite{srivastava2014dropout} with the drop rate $0.2$ between different layers. We use a special UNK token to replace the input lemmas, POS tags, and NER tags with a rate of $0.33$. Parameter optimization is performed with the Adam optimizer \cite{kingma2014adam} with $\beta_1=0.9$ and $\beta_2=0.999$. The same learning rate schedule of \cite{vaswani2017attention} is adopted in our experiments. We use early stopping on the development set for choosing the best model.

Following \newcite{lyu2018amr}, for word sense disambiguation, we simply use the most frequent sense in the training set, or \texttt{-01} if not presented. For wikification, we look-up in the training set for the most frequent one and default to ``-”.
	\begin{table}[t]
	\centering
	\small
	\begin{tabular}{c|c|c}
		\hline
		model component&  hyper-parameter& value\\
		\hline
		\multirow{4}{*}{char-level CNN} & number of filters & 256 \\
		& width of filters & 3 \\
		& char embedding size & 32 \\
		& final hidden size & 128 \\
		\hline
		\multirow{3}{*}{Transformer} & number of heads & 8 \\
		& hidden state size & 512 \\
		& feed-forward hidden size  & 1024 \\
		\hline
		\multirow{4}{*}{Sentence Encoder} &Transformer layers& 4\\
		&lemma embedding size& 200\\
		& POS tag embedding size & 32 \\
		& NER tag embedding size & 16 \\
		\hline
		\multirow{2}{*}{Graph Encoder} & Transformer layers & 1 \\
		&concept embedding size&300\\
		\hline
		Focus Selection & attention layers & 3 \\
		\hline
		Relation Identification & number of heads & 8 \\
		\hline
		Relation Classification & hidden state size & 100 \\
		\hline
	\end{tabular}
	\caption{Hyper-parameters settings. }
	\label{setting}
\end{table}
\end{document}